\documentclass{new_tlp}
\usepackage{mathptmx}
\usepackage{multirow}
\usepackage{listings}

  \title[Theory and Practice of Logic Programming]
        {A Physician Advisory System for Chronic Heart Failure Management Based on Knowledge Patterns
	}

  \author[Zhuo Chen, Kyle Marple, Elmer Salazar, Gopal Gupta, Lakshman Tamil]
		  {Zhuo Chen, Kyle Marple, Elmer Salazar, Gopal Gupta, Lakshman Tamil\\
         University of Texas at Dallas, Texas, U.S.A.\\
         \email{$\{$zxc130130,kbm072000,ees101020,gupta,laxman$\}$@utdallas.edu}}
       
\jdate{May 2016}
\pubyear{2016}
\pagerange{\pageref{firstpage}--\pageref{lastpage}}
\doi{}

\begin{document}
\bibliographystyle{acmtrans}

\nocite{*}

\label{firstpage}

\maketitle

  \begin{abstract}
    Management of chronic diseases such as heart failure, diabetes, and chronic obstructive pulmonary disease (COPD) is a major problem in health care. A standard approach that the medical community has devised to manage widely prevalent chronic diseases such as chronic heart failure (CHF) is to have a committee of experts develop guidelines that all physicians should follow.  These guidelines typically consist of a series of complex rules that make recommendations based on a patient's information.  Due to their complexity, often the guidelines are either ignored or not complied with at all, which can result in poor medical practices. It is not even clear whether it is humanly possible to follow these guidelines due to their length and complexity. In the case of CHF management, the guidelines run nearly 80 pages. In this paper we describe a physician-advisory system for CHF management that codes the entire set of clinical practice guidelines for CHF using answer set programming. Our approach is based on developing reasoning templates (that we call knowledge patterns) and using these patterns to systemically code the clinical guidelines for CHF as ASP rules. Use of the knowledge patterns greatly facilitates the development of our system.  Given a patient's medical information, our system generates a recommendation for treatment just as a human physician would, using the guidelines. Our system will work even in the presence of incomplete information. Our work makes two contributions: (i) it shows that highly complex guidelines can be successfully coded as ASP rules, and (ii) it develops a series of knowledge patterns that facilitate the coding of knowledge expressed in a natural language and that can be used for other application domains. This paper is under consideration for acceptance in TPLP.
  \end{abstract}


\section{Introduction}
 Chronic diseases are health conditions that can neither be prevented nor be cured but can only be managed. They have been the major consumer of health-care funds throughout the world. In America alone there are more than 133 million people---which is more than 40\% of the U.S. population---who suffer from one or more chronic diseases \cite{RefWorks:206}. In the U.S. they account for 81\% of hospital admissions, 91\% of prescriptions filled and 76\% of all physician visits \cite{RefWorks:208}. Though the list of chronic conditions is long, the top five conditions are: heart disease, cancers, stroke, chronic obstructive pulmonary disease (COPD) and diabetes. 
 In 2010, 68\% of the healthcare spending---more than trillion dollars---went towards the treatment of chronic diseases \cite{RefWorks:209}. The successful management of chronic diseases has two components: (i) self-management by the patients, and (ii) management by physicians while adhering to strict guidelines. The failure of either component will lead to the failure of the whole enterprise for the management of chronic diseases.
 
 We selected Chronic Heart Failure (CHF) as our first chronic disease to build tools to manage. Chronic Heart Failure is the inability of the heart to pump properly; consequently, not enough oxygen-rich blood can be supplied to all parts of the body. This causes congestion of blood in the lungs, abdomen, legs, etc., causing uneasiness while carrying out any kind of physical activity. Half of the people diagnosed with CHF die within five years. In U.S. alone there are 5.7 million people currently living with CHF \cite{RefWorks:193}. This disease affects all age groups including children. The cost of this disease to the economy is enormous. The statistics show that there are 11 million physician visits and 875,000 hospitalizations per year due to CHF. 25\% of patient with CHF are readmitted to a hospital or visit an emergency room within thirty days of treatment. This causes hospitals to lose money, as new regulations require them to bear the cost of readmission if it occurs within a period of thirty days. 
 
 Our work on chronic disease management started several years ago with the building of a tele\-medicine-based chronic disease management system designed to facilitate the self-management of chronic diseases by patients \cite{RefWorks:212,Savio-phd}. Our chronic care platform was put to test to see whether it can successfully prevent patients from going to an emergency room or being readmitted to a hospital within 30 days of discharge after the original CHF episode. This was done in collaboration with Texas Health Resources (THR), one of the largest hospital chains in the southwestern United States. Statistically, 25\% of discharged patients would be readmitted within 30 days. In our clinical study, 12 patients discharged from Texas Methodist Hospital in Cleburne, Texas after CHF episodes were chosen as subjects. (Note that though the number of patients used in the trial is not statistically significant to make any generalization but does prove the viability of our system). These patients were provided with our chronic care platform and peripheral devices such as a weighing scale, blood pressure meter, gluco-meter and pulse oximeter (these devices measure the vitals directly and are connected to our telemedicine device via Bluetooth). After some training by case nurses, the patients measured their vital signs periodically using these Bluetooth-enabled devices and answered questionnaires about their health. Using our platform, their vital sign data was automatically uploaded to a server where it was available for processing and viewing. Working remotely, the case nurse was able to manage the patients and prevent all of them from begin readmitted within 30 days \cite{RefWorks:213}. The telemedicine platform that we initially developed, and that we have licensed to a commercial company, only facilitated the transfer of information from patients residing remotely to the doctors and nurses providing care. 
 
 In this paper, we focus on the second component of CHF management, namely, a Physician Advisory System. This system assists physicians in adhering to the guidelines for managing CHF. The CHF management guidelines are published by the American College of Cardiology Foundation (ACCF) and the American Heart Association (AHA). The most recent version is the 2013 ACCF/AHA Guideline for the Management of Heart Failure \cite{RefWorks:190}. These guidelines were created by a committee of physicians based on thorough review of clinical evidence on heart failure management. They represent a consensus among the physicians on the appropriate treatment and management of heart failure \cite{RefWorks:191}. Though evidence-based guidelines should be the basis for all disease management \cite{RefWorks:207}, physicians' adherence to guidelines is very poor \cite{RefWorks:200}. The major reasons for the failure of guideline implementation are lack of awareness, lack of familiarity, lack of motivation and external barriers. For 78\% of clinical practice guidelines, more than 10\% of the physicians are not aware of their existence. Even when the guidelines are readily accessible, the physicians are not familiar enough with the guidelines to apply them correctly. In all the physician surveys conducted, the lack of familiarity was more common than the lack of awareness \cite{RefWorks:200}. 
 
 One of the reasons for the lack of familiarity is that the guidelines can be quite complex, as in the case of CHF management. For example, more than 100 variables have been associated with mortality and re-hospitalization related to heart failure. In the 2013 ACCF/AHA Guideline for the Management of Heart Failure, the variables range from simple information like age and sex to sophisticated data like the patterns in electrocardiogram and history of CHF-related symptoms and diseases. The rules for treatment recommendation in the guideline look like the following:
 
 \begin{quote}
 \narrower{\noindent ``Aldosterone receptor antagonists are recommended to reduce morbidity and mortality following an acute MI in patients who have LVEF of 40\% or less who develop symptoms of HF or who have a history of diabetes mellitus, unless contraindicated.''}
 \end{quote}

 \begin{quote}
 \narrower{\noindent ``In patients with a current or recent history of fluid retention, beta blockers should not be prescribed without diuretics.''}
 \end{quote}
 
 With more than 60 rules like the ones above, giving correct recommendations becomes an error-prone task for even the most experienced physicians \cite{RefWorks:199}.
 
To overcome the difficulties that physicians face in implementing the guidelines, we have developed a Physician Advisory System that automates the 2013 ACCF/AHA Guideline for the Management of Heart Failure. Our physician advisory system is able to give recommendations like a real human physician who is following the guidelines strictly, even under the condition of incomplete information about the patient. 

Our physician-advisory system for chronic heart failure management relies on answer set programming \cite{RefWorks:205,asp-paper} for coding the guidelines. The guideline rules are fairly complex and require the use of negation as failure, non-monotonic reasoning and reasoning with incomplete information. A fairly common situation in medicine is that a drug can only be recommended if its use is not contraindicated (i.e., the use of the drug will not have an adverse impact on that patient). Contraindication is naturally modeled via negation as failure. The ability of answer set programming to model defaults, exceptions, weak exceptions, preferences, etc., makes it ideally suited for coding these guidelines. Traditional techniques such as logic programming (Prolog) and production systems (OPS5), or traditional expert system styled approaches will result in a far more complex system due to the inability of these systems to model negation as failure effectively \cite{Baralbook}. For example, in Prolog, negation as failure is unsound for non-grounded goals, and if care is not taken, then use of negation as failure in Prolog programs can often lead to non-termination. Thus, coding our system in these formalisms would be a much more difficult and complex task. In contrast, the CHF guidelines can be coded in ASP very naturally (it took about 2 months to develop the first version of the system).
 
With the help of a telemedicine platform \cite{RefWorks:213} and electronic medical records (EMR) integrated with our ASP-based system, significant automation can be achieved. Besides boosting the physicians' compliance with the guidelines, the system can enable patients with CHF to manage their conditions with minimal supervision by doctors. The rest of this paper presents our ASP-based physician-advisory system as well as the knowledge patterns we have developed. We assume that the reader is familiar with ASP \cite{asp-paper}.

\section{Automating Chronic Heart Failure Management}

The 2013 ACCF/AHA Guideline for the Management of Heart Failure is intended to assist healthcare providers in clinical decision making by describing a range of generally acceptable approaches for the management of chronic heart failure. The guideline is based on four progressive stages of heart failure. Stage A includes patients at risk of heart failure who are asymptomatic and do not have structural heart disease. Stage B describes asymptomatic patients with structural heart diseases; it includes New York Heart Association (NYHA) class I, in which ordinary physical activity does not cause symptoms of heart failure. Stage C describes patients with structural heart disease who have prior or current symptoms of heart failure; it includes NYHA class I, II (slight limitation of physical activity), III (marked limitation of physical activity) and IV (unable to carry on any physical activity without symptoms of heart failure, or symptoms of heart failure at rest). Stage D describes patients with refractory heart failure who require specialized interventions; it includes NYHA class IV. Interventions at each stage are aimed at reducing risk factors (stage A), treating structural heart disease (stage B) and reducing morbidity and mortality (stages C and D) \cite{RefWorks:190}. 

The input to the system is a patient's information, including demographics, history, daily symptoms, risks and measurements, as well as ACCF/AHA stage and NYHA class. When queried for a treatment recommendation, our system is able to give recommendations according to the guideline just as a physician would. 

Our system is designed for running on top of the s(ASP) system, a goal-directed, predicate ASP system that can be thought of as Prolog extended with negation based on the stable model semantics \cite{RefWorks:204}. Because of the goal-directed nature of the system, only the particular treatments applicable to the patient are reported by the system. With minor changes, our system will also work with traditional SAT-based implementations such as CLASP \cite{clasp,clasp2}. However, these systems will compute the entire model, so if there are multiple treatments for a given condition, they will all be included in the answer set (these differences between goal-directed and SAT-based solvers are explained in \cite{galliwasp}). 

To implement the CHF guidelines in ASP, we first extensively studied the guidelines to extract reasoning templates. These templates can be thought of as general knowledge patterns. These patterns were next deployed to code the CHF guideline rules. Our research makes two major contributions: 

\begin{enumerate}

\item We develop a system that completely automates the entire set of guidelines for CHF management developed by the American College of Cardiology Foundation and American Heart Association. The system takes its input from (i) a patient's electronic health record that includes demographic information, test results, etc., and (ii) a telemedicine system that provides data about vital signs (heart rate, blood pressure, weight, etc.). It then uses this information to recommend a treatment. The s(ASP) system also supports abduction, thus our system can also be used for abductive reasoning: a physician can, for example, figure out the symptoms that a particular patient must have in order for a given treatment to work. 
\item We develop a set of general knowledge patterns that were used to realize our automated physician-advisory system and that can be helpful in translating rules expressed in a natural language into ASP for any application domain.  

\end{enumerate}

\section{A Physician-Advisory System for CHF Management}
\label{sec:A physician-advisory system for CHF management}

\subsection{System Description}

A large number of software systems have been designed to address CHF. However, none of them are designed to automatically advise physicians based on the ACCF/AHA guidelines. Chronic disease management systems designed thus far fall into seven categories \cite{RefWorks:194}: accessibility, care management, point-of-care functions, decision support, patient self-management, population management, and reporting. The automation of these functionalities is certainly helpful in assisting health care providers with managing patients with chronic conditions, however, none of them cover what we have realized: a physician advisory system that automates the application of clinical practice guidelines.
As mentioned earlier, our physician advisory system for CHF management codes all the knowledge in the 2013 ACCF/AHA Guideline for the Management of Heart Failure \cite{RefWorks:190} as an answer set program. Our system is able to recommend treatments just like a human physician who is strictly following these guidelines. Additionally, our system is able to recommend treatments even when a patient's information is incomplete. The input to our system is the patient's information which includes demographics, history, daily symptoms, known risk factors, measurements as well as ACCF/AHA stage and NYHA class \cite{RefWorks:190}. The inputs to our system are summarized in Table \ref{tab:input}. A physician uses our system by posing a query to it. Our system then processes the query by essentially simulating the thinking process of a CHF specialist (represented by the ACCF/AHA guideline).

\begin{table}
\caption{Input of the Physician-Advisory System for CHF Management}
\begin{minipage}{\textwidth}
\begin{tabular}{ll}

\hline\hline

\multirow{1}{*}{Demographics} 

& Gender; age; race \\

\hline
\multirow{1}{*}{Measurements} 
& Weight; creatinine; potassium; sinus rhythm; left bundle branch block; \\ 
& non-left bundle branch block; QRS duration; ejection fraction \\
& NYHA class; ACCF/AHA stage;\\
\hline

\multirow{1}{*}{Diseases and Symptoms} 
& Sleep apnea, acute coronary syndrome; myocardial infarction; obesity;\\
& diabetes; stroke; fluid retention; angioedema; ischemic attack; \\
& thromboembolism; elevated plasma natriuretic peptide level;   \\
& asymptomatic ischemic cardiomyopathy; lipid disorders; hypertension; \\
& atrial fibrillation; myocardial ischemia; coronary artery disease; \\
& dilated cardiomyopathy; acute profound hemodynamic compromise;  \\
& threatened end organ dysfunction; ischemic heart disease;angina; \\
& structural cardiac abnormalities; atrioventricular block; volume overload \\
\hline

\multirow{1}{*}{Miscellany} 
& Expectation of survival;pregnancy; history of cardiovascular hospitalization;\\
& history of standard neurohumoral antagonist therapy; \\
& risk of cardioembolic stroke; eligibility of significant ventricular pacing; \\
& eligibility of mechanical circulatory support;   \\
& dependence of continuous parenteral inotropic; ischemic etiology of HF; \\
& requirement of ventricular pacing \\
\hline\hline
\end{tabular}
\vspace{-2\baselineskip}
\label{tab:input}
\end{minipage}
\end{table}

The physician-advisory system for CHF management has two major components, the rules database and the fact table. The rules database covers all the knowledge in 2013 ACCF/AHA Guideline for the Management of Heart Failure \cite{RefWorks:190}. The fact table contains the relevant information of the patient with heart failure. The fact table is derived from a patient's electronic health record and from a telemedicine system used to measure vital signs. The patient information consists mainly of: 5 pieces of demographics information, 8 measurements and 25 types of HF-related diseases and symptoms. Treatment recommendations returned by the system may include: 11 pharmaceutical treatments, 9 management objectives, and 4 device/surgery therapies. Table \ref{tab:output} displays the outputs produced by our system.

\begin{table}
\caption{Output of the Physician-Advisory System for CHF Management}
\begin{minipage}{\textwidth}
\begin{tabular}{ll}

\hline\hline
		
\multirow{1}{*}{Pharmaceutical Treatments} 
& ACE inhibitors; ARBs; Beta blockers; statin; diuretics; \\
& aldosterone receptor antagonists; hydralazine and isosorbide dinitrate; \\
& digoxin; anticoagulations; Omege-3 fatty acids; inotropes; \\
		
\hline
\multirow{1}{*}{Management Objectives} 
& Systolic blood pressure control; diastolic blood pressure control; \\
& obesity control; diabetes control; tobacco avoidance; \\
& cardiotoxic agents avoidance; atrial fibrillation control; water restriction; \\
& sodium restriction; \\ 

\hline
		
\multirow{1}{*}{Device/Surgery Therapies} 
& Implantable cardioverter-defibrillator; \\
& cardiac resynchronization therapy; \\
& mechanical circulatory support; coronary revascularization \\

\hline\hline
\end{tabular}
\vspace{-2\baselineskip}
\label{tab:output}
\end{minipage}
\end{table}

There are some sixty odd rules in the 2013 ACCF/AHA Guideline for the Management of Heart Failure. All of these rules are coded in ASP to run on the s(ASP) system \cite{RefWorks:204}. For example, take one rule from ACCF/AHA Stage B \cite{RefWorks:190}:

\begin{quote}
\narrower{\noindent ``In all patients with a recent or remote history of MI or ACS and reduced EF, evidence-based beta blockers should be used to reduce mortality.''}
\end{quote}

\noindent In s(ASP), this rule will be coded as shown below. Note that Prolog conventions are followed (variables begin with an upper case letter).

\begin{verbatim}
recommendation(beta_blockers, class_1):- accf_stage(b), 
    history_of_mi_or_acs, measurement(lvef, Data),
    reduced_ef(Data), not contraindication(beta_blockers).
\end{verbatim}

Due to using ASP as the modeling formalism, our system is able to handle complex scenarios. To illustrate this, consider a situation in which we have a patient who has heart failure with reduced ejection fraction (HFrEF) and is in ACCF/AHA stage C. According to the guideline \cite{RefWorks:190} our system would recommend the use of beta blockers. However, if we add the information that the patient has a history of fluid retention, then our system would try to add diuretics due to the following rule: ``In patients with a current or recent history of fluid retention, beta blockers should not be prescribed without diuretics'' \cite{RefWorks:190}. However, if diuretics are contraindicated for any reason, then our system will not recommend beta blockers either. 
Our software system has been tested in-house and outputs found to be consistent with our interpretation of the rules. More extensive testing and a clinical trial are planned in the near future in collaboration with medical doctors who are CHF specialists. 

Our system can also perform abductive reasoning thanks to the s(ASP) system's support for abduction \cite{RefWorks:204}. Abductive reasoning is a form of logical inference where one attempts to augment a theory with sufficient information to explain an observation (the augmentations come from a set of predicates that are declared as {\it abducibles}). To illustrate, consider the following two rules in the ACCF/AHA guideline \cite{RefWorks:190}:

\begin{itemize}
	\item Combination of hydralazine \& isosorbide dinitrate is recommended to reduce morbidity \& mortality for patients self-described as African Americans with NYHA class III-IV HFrEF receiving optimal therapy with ACE inhibitors \& and beta blockers, unless contraindicated.
	
	\item Combination of hydralazine \& isosorbide dinitrate should not be used for the treatment of HFrEF in patients who have no prior use of standard neurohormonal antagonist therapy.
\end{itemize}

Suppose we have an African American patient who is suffering from NYHA class III HFrEF, but that is all we know about the patient. Since a hydralazine and isosorbide dinitrate combination is highly effective in reducing the mortality of African Americans with HFrEF, the physician might pose the following query:

 {\tt ?-recommendation((hydralazine\_and\_isosorbide\_dinitrate), class\_1)}

\noindent to the s(ASP) system.  The system would return the results shown in figure \ref{fig:abduction result}.
\begin{figure}
	\figrule
\{accf\_stage(c), hf\_with\_reduced\_ef, history(standard\_neurohormonal\_antagonist\_therapy), nyha\_class(3), nyha\_class\_3\_to\_4, race(african\_american), recommendation(hydralazine\_and\_isosorbide\_dinitrate,class\_1), not contraindication(hydralazine\_and\_isosorbide\_dinitrate)\}
	
	\caption{Result of abductive reasoning in physician-advisory system for CHF management}\label{fig:abduction result}
	\figrule
\end{figure}

Note that the system abduced two things: (i) a ``history of standard neurohormonal antagonist therapy", and (ii) the absence of ``contraindication of hydralazine and isosorbide dinitrate". This means in order for us to recommend hydralazine and isosorbide dinitrate to the patient, they must have received standard neurohormonal antagonist therapy before. Otherwise, hydralazine and isosorbide dinitrate would be contraindicated.

\subsection{Knowledge patterns in the guidelines for the management of heart failure}
\label{sec:Knowledge patterns}

The ACCF/AHA guidelines are written in English and are quite complex. Our task was to code these guidelines in ASP. To simplify our task, we developed reasoning templates that we call knowledge patterns. These knowledge patterns are quite general and serve as solid building blocks for systematically translating the specifications written in English to ASP. While developing these knowledge patterns and coding them in ASP, certain facts had to be noted: 
	(i) Multiple rules can lead to the recommendation of a treatment;
	(ii) Multiple rules can lead to contraindication of a treatment;
	(iii) A treatment cannot be recommended if at least one contraindication for that treatment is present; and,
	(iv) A given treatment recommendation can impact the recommendation and/or contraindication of other treatments.

Next, we present the most salient knowledge patterns that we have developed. Many of these patterns are straightforward, however, some of them, such as the concomitant choice rule, are intricate. We present these patterns at a high level and ignore non-essential details.

\noindent{\bf 1. Aggressive Reasoning:}
The aggressive reasoning pattern can be stated as ``take an action (e.g., recommend treatment) if there is a reason; no evidence of danger means there is no danger in taking that action''. The aggressive reasoning pattern is coded as follows: 

		\begin{verbatim}
		recommendation(Choice) :- preconditions(Choice), 
		    not contraindication(Choice).
		contraindication(Choice) :- dangers(Choice).
		\end{verbatim}

The code above makes use of negation as failure. If the contraindication of a choice cannot be proved, and the conditions for making the choice hold, then that choice is recommended. An example of this knowledge pattern can be found in \cite{RefWorks:190}: ``Digoxin can be beneficial in patients with HFrEF, unless contraindicated, to decrease hospitalizations for HF.'' It is realized in our system as shown below.


		\begin{verbatim}
		recommendation(digoxin, class_2a) :- not contraindication(digoxin),
		    accf_stage(c), hf_with_reduced_ef.
		contraindication(digoxin) :- evidence(atrioventricular_block).
		\end{verbatim}

\noindent{\bf 2. Conservative Reasoning:}
This reasoning pattern is stated as ``A reason for a recommendation is not enough; evidence that the recommendation is not harmful must be available''.

The conservative reasoning pattern is coded as follows: 

		\begin{verbatim}
		recommendation(Choice) :- preconditions(Choice), 
		    not contraindication(Choice).
		contraindication(Choice) :- not -dangers(Choice).
		\end{verbatim}

This coding pattern requires evidence of the absence of danger. Without such evidence, the choice would be considered contraindicated. Note that the ``-'' operator indicates classical negation.
An example of this knowledge pattern can be found in \cite{RefWorks:190}: ``In patients with structural cardiac abnormalities, including LV hypertrophy, in the absence of a history of MI or ACS, blood pressure should be controlled in accordance with clinical practice guidelines for hypertension to prevent symptomatic HF.'' It is realized in our system as shown below.


		\begin{verbatim}
		recommendation(blood_pressure_control, class_1):-
		    accf_stage(b), diagnosis(structural_cardiac_abnormalities),
		    not contraindication(blood_pressure_control).
		contraindication(blood_pressure_control):- not -history(mi).
		contraindication(blood_pressure_control):- not -history(acs).
		\end{verbatim}

\noindent{\bf 3. Anti-recommendation:}
The anti-recommendation pattern is stated as ``a choice can be prohibited if evidence of danger can be found''.

The coding pattern for the anti-recommendation is coded as follows: 

		\begin{verbatim}
		contraindication(choice) :- dangers(Choice).
		\end{verbatim}

The code above specifies the conditions under which a choice will be ruled out (i.e., contraindicated). Note that for a choice to be made, both aggressive reasoning and conservative reasoning require that the contraindication of the choice be false. 
An example of this knowledge pattern can be found in \cite{RefWorks:190}: ``Anticoagulation is not recommended in patients with chronic HFrEF without AF, a prior thromboembolic event, or a cardioembolic source.'' It is realized in our system as shown below.


		\begin{verbatim}
		contraindication(anticoagulation) :- not cardioembolic_source,
		    not diagnosis(af), not history(thromboembolism),
		    hf_with_reduced_ef.
		\end{verbatim}

\noindent{\bf 4. Preference:}
The preference pattern is stated as ``use the second-line choice when the first-line choice is not available''. The preference pattern is coded as follows: 

		\begin{verbatim}
		recommendation(First_choice) :- conditions_for_both_choices, 
		    not contraindication(First_choice).
		recommendation(Second_choice) :- conditions_for_both_choices, 
		    contraindication(First_choice), 
		    not contraindication(Second_choice).
		\end{verbatim}

This code chooses the treatment recommendation in the second choice only when the conditions are satisfied, the first choice is contraindicated, and there is no evidence preventing the use of second choice.
An example of this knowledge pattern can be found in \cite{RefWorks:190}: ``ARBs are recommended in patients with HFrEF with current or prior symptoms who are ACE inhibitor intolerant, unless contraindicated, to reduce morbidity and mortality.'' It is realized in our system as shown below.


		\begin{verbatim}
		recommendation(ace_inhibitors, class_1) :- 
		    not contraindication(ace_inhibitors),
		    accf_stage(c), hf_with_reduced_ef.
		recommendation(arbs, class_1) :- contraindication(ace_inhibitors),
		    not contraindication(arbs), 
		    not taboo_choice(arbs),
		    accf_stage(c), hf_with_reduced_ef.
		\end{verbatim}

\noindent{\bf 5. Concomitant Choice:}
The concomitant choice pattern is stated as ``if a choice is made, some other choices are automatically in effect unless they are prohibited.'' The concomitant pattern is coded as shown below.

		\begin{verbatim}
		recommendation(Trigger_choice) :- preconditions(Trigger_choice), 
		    not contraindication(Trigger_choice), 
		    not skip_concomitant_choice(Trigger_choice).
		skip_concomitant_choice(Trigger_choice) :- 
		    not recommendation(Concomitant_choice), 
		    not contraindication(Concomitant_choice).
		recommendation(Concomitant_choice) :- 
		    recommendation(Trigger_choice),
		    not contraindication(Concomitant_choice).
		\end{verbatim}

The above code makes sure that a concomitant choice appears in all stable models containing the trigger choice, provided the concomitant choice is not prohibited. The trigger choice is always recommended along with the concomitant choice unless the concomitant choice is contraindicated. An example of this knowledge pattern can be found in \cite{RefWorks:190}: ``Diuretics should generally be combined with an ACE inhibitor, beta blocker, and aldosterone antagonist. Few patients with HF will be able to maintain target weight without the use of diuretics.'' It is realized in our system as shown below. To save space, we list only the code for ACE inhibitors.


		\begin{verbatim}
		recommendation(ace_inhibitors, class_1) :- accf_stage(c),
		    not skip_concomitant_choice(ace_inhibitors),
		    not contraindication(ace_inhibitors), hf_with_reduced_ef.
		skip_concomitant_treatment(ace_inhibitors) :-
		    hf_with_reduced_ef, not recommendation(diuretics, class_1),
		    not contraindication(diuretics).
		recommendation(diuretics, class_1) :-
		    hf_with_reduced_ef, not contraindication(diuretics),
		    recommendation(ace_inhibitors, class_1).
		\end{verbatim}

\noindent{\bf 6. Indispensable Choice:}
The indispensable choice pattern is stated as ``if a choice is made, some other choices must also be made; if those choices can't be made, then the first choice is revoked''. Note that choosing “Trigger\_choice” forces “Indispensable\_choice”. The indispensable choice pattern is coded as shown below:

		\begin{verbatim}
		recommendation(Trigger_choice) :- preconditions(Trigger_choice),
		    not contraindication(Trigger_choice)，
		    not absent_indispensable_choice(Trigger_choice).
		absent_indispensable_choice(Trigger_choice) :- 
		    not recommendation(Indispensable_choice).
		recommendation(Indispensable_choice) :- recommendation(Trigger_choice), 
		    not contraindication(Indispensable_choice).
		\end{verbatim}

The above code makes sure that the trigger choice will always appear with the indispensable choice. If for some reason the indispensable choice cannot be made, then the trigger choice cannot be made either. 
An example of this knowledge pattern can be found in \cite{RefWorks:190}: ``In patients with a current or recent history of fluid retention, beta blockers should not be prescribed without diuretics''. It is realized in our system as shown below.


		\begin{verbatim}
		recommendation(beta_blockers, class_1) :-
		    not skip_concomitant_choice(beta_blockers),
		    not absent_indispensable_choice(beta_blockers),
		    not contraindication(beta_blockers), accf_stage(c), hf_with_reduced_ef.
		absent_indispensable_choice(beta_blockers) :-
		    not recommendation(diuretics, class_1), hf_with_reduced_ef, 
		    accf_stage(c), current_or_recent_history_of_fluid_retention.
		recommendation(diuretics, class_1) :-
		    recommendation(beta_blockers, class_1),
		    not contraindication(diuretics), accf_stage(c), hf_with_reduced_ef,
		    current_or_recent_history_of_fluid_retention.
		\end{verbatim}

\noindent{\bf 7. Incompatible Choice:}
The incompatibility pattern is stated as ``some choices cannot be in effect at the same time''. The incompatible choice pattern is coded as shown below:

		\begin{verbatim}
		
		taboo_choice(Choice_1) :-             recommendation(Choice_1) :- 
		    recommendation(Choice_2),             conditions_for_choice_1,
		    ...,                                  not contraindication(Choice_1),
		    recommendation(Choice_n).             not taboo_choice(Choice_1). 		    
		taboo_choice(Choice_2) :-             recommendation(Choice_2) :- 
		    recommendation(Choice_1),             conditions_for_choice_2,
		    recommendation(Choice_3),             not contraindication(Choice_2),
		    ....                                  not taboo_choice(Choice_2).
		    recommendation(Choice_n).
	    ...                                   ...
		taboo_choice(Choice_n) :-             recommendation(Choice_n) :-
		    recommendation(Choice_1),             conditions_for_choice_n,
		    recommendation(Choice_2),             not contraindication(Choice_n),
		    ....                                  not taboo_choice(Choice_n).
		    recommendation(Choice_n-1).			   	
		\end{verbatim}

The above code makes sure that incompatible choices will not be made together. Note that we did not use a simple constraint to implement this pattern. A constraint would kill all stable models if each of the choices in question can be made. Our implementation, on the other hand, will produce partial answer sets supporting the query, thanks to the goal-driven mechanism of s(ASP) \cite{RefWorks:204}.
An example of this knowledge pattern can be found in \cite{RefWorks:190}: ``Routine combined use of an ACE inhibitor, ARB, and aldosterone antagonist is potentially harmful for patients with HFrEF.'' It is realized in our system as shown below.


		\begin{verbatim}
		taboo_choice(ace_inhibitors) :- hf_with_reduced_ef, 
		    recommendation(arbs, class_1), 
		    recommendation(aldosterone_antagonist, class_1).
		taboo_choice(arbs) :- hf_with_reduced_ef, 
		    recommendation(ace_inhibitors, class_1), 
		    recommendation(aldosterone_antagonist, class_1).
		taboo_choice(aldosterone_antagonist) :- hf_with_reduced_ef, 
		    recommendation(arbs, class_1), recommendation(ace_inhibitors, class_1).
		recommendation(ace_inhibitors, class_1) :- accf_stage(c),
		    hf_with_reduced_ef, not skip_concomitant_choice(ace_inhibitors),
		    not taboo_choice(ace_inhibitors), not contraindication(ace_inhibitors).   
		recommendation(arbs, class_1) :- contraindication(ace_inhibitors),
		    not contraindication(arbs), not taboo_choice(arbs),
		    accf_stage(c), hf_with_reduced_ef.
		recommendation(aldosterone_antagonist, class_1) :-
		    conditions_for_aldosterone_antagonist_class_1,
		    not skip_concomitant_choice(aldosterone_antagonist),
		    not contraindication(aldosterone_antagonist),
		    not taboo_choice(aldosterone_antagonist).
		\end{verbatim}

\section{Results and discussion}

Our system has been tested in-house and has shown accurate results that are 
compatible with what a physician following the guidelines would conclude.  
A clinical trial is planned. 

To illustrate how our system works, consider a female heart failure 
patient who is in ACCF/AHA stage C, belongs to NYHA class 3 and has been 
diagnosed as myocardial ischemia, atrial fibrillation, coronary artery disease. 
She also suffers from sleep apnea, fluid retention and hypertension. Her left 
ventricular ejection fraction (LVEF) is 35\%. There is evidence that she has 
ischemic etiology of heart failure. Her electrocardiogram (ECG) has sinus 
rhythm and a left bundle branch block (LBBB) pattern with a QRS duration of 
180ms.The blood test says her creatinine is 1.8 mg/dL and potassium is 4.9 
mEg/L. She has a history of stroke. It has been 40 days since the acute 
myocardial infarction happened to her. Her doctor assessed that her expectation 
of survival is about 3 years.

\begin{figure}
	\figrule
	\begin{center}
		\begin{verbatim}
		%doctor's assessments                    %history of the patient
		accf_stage(c).                           diagnosis(myocardial_ischemia).
		nyha_class(3).                           diagnosis(atrial_fibrillation).
		expectation_of_survival(3).              diagnosis(coronary_artery_disease).
		                                         diagnosis(hypertension).
		%demographics of the patient             evidence(ischemic_etiology_of_hf).
		gender(female).                          evidence(sleep_apnea).
		age(78).                                 evidence(fluid_retention).
		                                         history(mi, recent).
		%measurements from the lab               history(stroke).
		hf_with_reduced_ef.                      history(cardiovascular_hospitalization).
		measurement(creatinine, 1.8).            post_mi(40).
		measurement(potassium, 4.9).
		measurement(lvef, 0.35).
		measurement(lbbb, 180).
		measurement(sinus_rhythm).	
		\end{verbatim}
	\end{center}
	\caption{Representation of a patient's information in physician-advisory system for CHF management}\label{fig:Representation of a patient's information 
		in physician-advisory system for CHF management}
	\figrule
\end{figure}

The patient's information derived from her electronic health record is coded
as the facts shown in Figure \ref{fig:Representation of a patient's information 
in physician-advisory system for CHF management}.
There are multiple treatments for this patient. Figure \ref{fig:output} shows some
of the treatment recommendations our system 
infers once we give the query {\tt recommendation(Treatment, Class)}. 
Each treatment
recommendation (represented as a partial answer set)
contains all of the predicates that must hold in order for the query to be successful.
For instance, consider the recommendation of ace inhibitors as a treatment option (answer \#2). 
Ace inhibitors are recommended because the patient is in ACCF/AHA Stage C, per the doctor's
assessment, and has heart failure with reduced ejection fraction condition.
Proof of contraindication for ace inhibitors is absent
as the patient does not have a history of angioedema ({\tt not history(angiodema)}) 
and is not pregnant ({\tt not pregnancy}). The system also 
gives us the concomitant treatments for ace inhibitors, namely, beta blockers 
and diuretics. It is worth mentioning that we used the aggressive reasoning 
pattern (see section \ref {sec:Knowledge patterns}) when coding the rules of 
ace inhibitors. 
Had we adopted the conservative reasoning pattern, ace inhibitors 
would not have been recommended unless we explicitly asserted
{\tt -history(angioedema)} and {\tt -pregnancy} in the patient's information (a definitive
proof of the latter can be derived from patient's age (78)).

Given that there may be multiple treatment options for a particular patient, the choice of a particular treatment
will depend on the physician's preference. Rules that capture a physician's or a nurse's preference can also be
coded as answer set programs in our system.

While our testing indicates that the system works well and the results produced
are consistent with what a physician may recommend, if they were to exactly 
follow the guidelines, a clinical trial is needed to truly validate our 
system, and is indeed planned. As mentioned earlier, our system can be used for abductive reasoning as well. Running the system in the abductive mode can allow a physician to try out 
what-if scenarios and to make sure that all the pre-conditions required for
treatment are met.

\begin{figure}
  \figrule
	\{ accf\_stage(c), recommendation(sodium\_restriction,class\_2a), not 
	contraindication(sodium\_restriction) \}
	Treatment = sodium\_restriction,
	Class = class\_2a
	
	\bigskip
	\{ accf\_stage(c), hf\_with\_reduced\_ef, 
	recommendation(ace\_inhibitors,class\_1), 
	recommendation(beta\_blockers,class\_1), recommendation(diuretics,class\_1), 
	not contraindication(ace\_inhibitors), not contraindication(beta\_blockers), 
	not contraindication(diuretics), not history(angioedema), not 
	history(angioedema,recent), not history(angioedema,remote), not pregnancy \}
	Treatment = ace\_inhibitors,
	Class = class\_1
	\medskip
	
%
	
%
	
	\caption{Output of the physician-advisory system for CHF management}
	\label{fig:output}
\figrule
\end{figure}

\section{Conclusion and future work}

In this paper we report on our work on developing a ASP-based physician advisory system for managing CHF using a telemedicine platform. The system automates
the rules laid out in the 2013 ACCF/AHA Guide for the Management of Heart Failure.  
It is able to take a patient's data as input and produce 
treatment recommendations that strictly adhere to the guidelines. It can also
be used by a physician to abduce symptoms and other conditions that must be
met by a given treatment recommendation.

Our approach to developing the system was 
based on identifying knowledge patterns and using them as building blocks for 
constructing the ASP code. 
There are many ways to further extend our work that we plan to pursue in the future: 

\begin{itemize}
	\item Extending the system for comorbidities: We would like our
	system to handle comorbidities of heart 
	failure \cite{RefWorks:194}. A typical CHF patients
	suffers from other chronic ailments as well, i.e., CHF generally never occurs by itself.
	
	\item Performing clinical trials: our system has been tested in-house,
	however, we plan to compare the recommendations given by our system to the prescriptions by human 
	cardiologists in a formal clinical trial to validate the effectiveness of our 
	system. 
	\item Integrating with EMRs and a Telemedicine Platform: Future work
	would include integrating our system with our telemedicine platform so that the input 
	comes directly from the electronic
	medical record while vital signs are directly obtained from the patient
	through our telemedicine hardware and software \cite{RefWorks:212,Savio-phd}. A 
	user-friendly GUI will also be designed to make the system more usable. 
	
	
	
	\item Adding justification to recommendations given by our system: 
	Although the rationale behind a recommendation is shown in the partial answer 
	set, it is hard to decipher it. We plan to augment s(ASP) 
	\cite{RefWorks:204} so that reasonably detailed justifications for a query are 
	printed in a human-readable form. 
	\item Formal Analysis: Conducting research to formally establish the correctness of our system. 
	\end{itemize}

\noindent {\bf Acknowledgment:} Thanks to Dr. Michael Skinner for discussion. This research is supported by
NSF (Grant No. 1423419) and the Texas Medical Research Collaborative.

\end{document}